**Deep Learning in Computational Biology: Advancements, Challenges, and Future Outlook**

**Suresh Kumar*, Dhanyashri Guruparan, Pavithren Aaron, Philemon Telajan, Kavinesh Mahadevan, Dinesh Davagandhi, Ong Xin Yue**

Faculty of Health and Life Sciences, Management and Science University, Seksyen 13, 40100, Shah Alam, Selangor, Malaysia

*Corresponding author: Tel: +60-14-2734893, Fax: +60-35-5112848, Email address: sureshkumar@msu.edu.my

**ORCID:**

Suresh Kumar: https://orcid.org/0000-0001-5682-0938

Dhanyashri Guruparan: https://orcid.org/0009-0007-0326-4609

Pavithren Aaron: https://orcid.org/0009-0001-4336-2314

Philemon Telajan: https://orcid.org/0009-0000-7611-6892

Kavinesh Mahadevan: https://orcid.org/0009-0003-4613-6122

Dinesh Davagandhi: https://orcid.org/0009-0008-2304-511X

Ong Xin Yue: https://orcid.org/0009-0005-6822-4670

## ABSTRACT

Deep learning has become a powerful tool in computational biology, revolutionising the analysis and interpretation of biological data over time. In our article review, we delve into various aspects of deep learning in computational biology. Specifically, we examine its history, advantages, and challenges. Our focus is on two primary applications: DNA sequence classification and prediction, as well as protein structure prediction from sequence data. Additionally, we provide insights into the outlook for this field. The history of deep learning can be traced back to the development of "threshold logic," which is a combination of algorithms and mathematics used to simulate cognitive processes. This breakthrough allowed deep learning to extract complex patterns from large-scale biological datasets. Deep learning offers several benefits, including the ability to accurately diagnose diseases, discover new drugs, and provide precision medicine. It also excels at managing large and complex datasets while improving generalisation. To fully harness the potential of deep learning in computational biology, it is crucial to address the challenges that come with it. These challenges include the requirement for large, labelled datasets and the interpretability of deep learning models. The use of deep learning in the analysis of DNA sequences has brought about a significant transformation in the detection of genomic variants and the analysis of gene expression. This has greatly contributed to the advancement of personalised medicine and drug discovery. Convolutional neural networks (CNNs) have been shown to be highly accurate in predicting genetic variations and gene expression levels. Deep learning techniques are used for analysing epigenetic data, including DNA methylation and histone modifications. This provides valuable insights into metabolic conditions and gene regulation. The field of protein structure prediction has been significantly impacted by deep learning, which has enabled accurate determination of the three-dimensional shape of proteins and prediction of their interactions. The future of deep learning in computational biology looks promising. With the development of advanced deep learning models and interpretation techniques, there is potential to overcome current challenges and further our understanding of biological systems.

## 1.0 INTRODUCTION

### 1.1 Deep Learning in Computational Biology

The field of computational biology involves integrating advanced computing techniques and biological research to gain a deeper understanding of the intricate nature of living systems, which are continuously progressing. This requires utilising data analysis and theoretical techniques, along with mathematical modelling and computational simulation, to investigate systems related to behaviour, social interaction, and biology. (1). The origins of machine learning in this area can be traced back to utilising neural network applications for scrutinising gene expression data during the 1990s. However, considerable advancements were made only after devising deep learning algorithms like artificial neural networks, especially convolutional (CNNs) or recurrent (RNNs) ones, that aimed at identifying complex patterns within intricate datasets and generating dependable predictions (2).

The DeepBind algorithm, created by researchers from Harvard and MIT in 2015, was a pioneering achievement of deep learning applied to computational biology. By utilising a complex neural network model, it could successfully pinpoint RNA-binding protein binding sites (3), thus uncovering hitherto unknown regulatory elements in the genome. As time has gone on, scientists have been increasingly turning towards deep learning as an effective tool for tackling various biological challenges, be they predicting protein structures or identifying genetic mutations linked to disease causation. For instance, AlphaFold is one such system developed by experts at DeepMind that harnesses advanced neural networks to deliver remarkably accurate insights into proteins' three-dimensional structure; this breakthrough has technically opened new frontiers within the structural biology realm.The use of deep learning in computational biology has resulted in noteworthy breakthroughs spanning diverse niches, like genomics, medical diagnosis, and drug discovery. To elaborate further, the application of deep learning techniques has proven beneficial in scrutinising genomic data with precision, thus detecting genetic defects causing diseases and leading scientists to devise customised treatments accordingly. Additionally, this technology is useful for anticipating functional pathways for newly identified drugs, thereby making it effortless and economical to pinpoint potential targets quickly rather than conducting trial-and-error experiments incessantly. Computational biologists can delve into immense quantities of genomic, proteomic, and other biological data using deep learning to uncover underlying patterns. This method is especially suitable for scrutinising voluminous and intricate datasets since it detects subtle patterns that conventional statistical approaches might overlook, allowing a more profound comprehension of biological systems.

Grand View Research predicts that the global market for deep learning in life sciences will climb to USD 34.83 billion by 2021, boasting an impressive compound annual growth rate (CAGR) of 34.3% from years past until then. This exponential rise is attributed primarily to amplified expeditions towards personalised medicine and mounting pressure to enhance drug discovery efficiency. (4)Although deep learning boasts promising advantages in computational biology, it also poses considerable hurdles. One of these obstacles is the significant amount of training data required by deep learning algorithms to yield optimal accuracy, a challenge that some areas of biology struggle with (3). Additionally, comprehending and deciphering the results produced by these algorithms can be difficult since they detect intricate and subtle patterns that may not align well with conventional biological models.

Despite the presence of obstacles to surmount, deep learning technology has the potential to revolutionise computational biology by enhancing our comprehension of biological systems and elevating healthcare outcomes. This article delves into how deep learning is utilised in computational biology and examines its associated difficulties as well as prospects. Furthermore, this article also emphasises noteworthy developments within the field while considering their possible impacts on personalised medicine and drug discovery in the upcoming years.

## 1.2 Brief History and Evolution of Deep Learning

The term "deep learning" was first introduced to the machine learning community by Rina Dechter in 1986 and to artificial neural networks by Igor Aizenberg and colleagues in 2000, specifically in relation to Boolean threshold neurons (5). The roots of deep learning can be traced back to 1943, when Walter Pitts and Warren McCulloch created a computer model inspired by the neural networks of the human brain. They devised "threshold logic," a combination of algorithms and mathematics, to simulate cognitive processes. Since then, deep learning has undergone continuous development, with only two notable interruptions corresponding to the periods known as the Artificial Intelligence Winters (5). The history and evolution of deep learning are shown in **Table 1.** Warren McCulloch and Walter Pitts laid the groundwork for neural networks by creating a computational model called threshold logic, which employed mathematical principles and algorithms (6). In 1958, Frank Rosenblatt introduced the perceptron, a two-layer neural network designed for pattern recognition using basic addition and subtraction operations. Rosenblatt also proposed the concept of additional layers, although their practical implementation did not occur until 1975.

In 1980, Kunihiko Fukushima introduced the Neoconitron, an artificial neural network with a hierarchical and multilayered structure that proved useful for tasks such as handwriting and pattern recognition. By 1989, researchers had developed algorithms for deep neural networks, but their training times were impractical, often taking several days to complete. In 1992, Juyang Weng presented Cresceptron, an automated method for recognising 3D objects in cluttered scenes.

During the mid-2000s, the term "deep learning" gained popularity following a significant paper by Geoffrey Hinton and Ruslan Salakhutdinov. Their research showcased the effectiveness of training neural networks with multiple layers by gradually training each layer separately. In 2009, at the NIPS Workshop on Deep Learning for Speech Recognition, it was revealed that pre-training neural networks could be omitted when working with extensive datasets, leading to notable reductions in error rates. By 2012, artificial pattern recognition algorithms had achieved human-level performance on specific tasks, and Google's deep learning algorithm garnered attention for its ability to identify the features of cats.

In 2014, Google made a noteworthy acquisition by purchasing DeepMind, an AI startup based in the UK, for £400 million. Subsequently, in 2015, Facebook integrated deep learning technology known as DeepFace into its operations, enabling automatic tagging and identification of individuals in photographs. DeepFace utilised deep networks comprising 120 million parameters, resulting in remarkable face recognition capabilities. Finally, in 2016, Google DeepMind's AlphaGo algorithm achieved mastery in the intricate game of Go, defeating professional player Lee Sedol in a widely publicised tournament held in Seoul.

### 1.3 Deep Learning Algorithm

Artificial neural networks are employed by deep learning to execute complex computations on vast quantities of data. These networks, also referred to as ANNs, consist of layers made up of interconnected neurons that collaborate in processing and acquiring knowledge from incoming data inputs.

Deep learning techniques require data to be processed through multiple layers of neural network algorithms. Each layer simplifies the representation of the data before it is passed on to the next one. A deep neural network that is fully connected comprises an input layer and several hidden layers linked consecutively. Every neuron obtains its input either from the previous-layer neurons or directly from the input layer. Consequently, a one-neuron output advances to become another group of inputs for next-layer neurons, eventually reaching the final stage where it becomes the ultimate output that emerges out of this network. Through a sequence of nonlinear transformations, the neural network's layers enable it to acquire intricate features from input data and thereby learn complex representations (7).

A variety of algorithms are used in these models, each with specific strengths for certain types of tasks. Radial Function Networks, Multilayer Perceptrons, and Self-Organising Maps Convolutional neural networks (CNNs), long-short-term memory networks (LSTMs), and recurrent neural networks (RNNs) are some examples of deep learning algorithms. CNNs, LSTMs, and RNNs are the three top-performing algorithms considered highly effective in tackling complex problems in deep learning workloads.

## 2.0 ADVANTAGES AND CHALLENGES OF USING DEEP LEARNING IN COMPUTATIONAL BIOLOGY.

The rapid advancements in genomics and imaging technologies have generated vast amounts of molecular and cellular profiling data from hundreds of sources worldwide. This surge in biological data volume and collection speed could present a challenge to traditional analysis approaches (8). The accumulation of biomedical data has attracted considerable attention from both industries and academia, highlighting the potential for applications in biological and healthcare research (9). To extract valuable insights from big data in bioinformatics, machine learning has emerged as a widely adopted and successful methodology. Machine learning algorithms will leverage training data to uncover underlying patterns, construct models, and make predictions based on the most suitable model (10). The application of deep learning to computational biology has the potential to revolutionise biology and medicine, but it also comes with its own set of challenges. A perspective on the use of deep learning in computational biology from a bird's eye view is shown in **Figure 1**.

### 2.1 Advantages of Using Deep Learning

Deep learning technology could catalyse disease diagnosis and prediction. According to Ching et al. (11), deep learning technology could be utilised to develop diagnostic tools by providing more meaningful and data-driven approaches that should be able to recognise and identify pathological samples with accuracy. They also suggested that, with the help of deep learning technologies, it can be used to screen large datasets in a matter of time, in addition to cutting unnecessary costs for drug discovery applications by identifying drug targets from large databases and the interactions or predicting the drug response. On top of that, the application of deep learning for drug repositioning based on transcriptomic data, which is used for describing gene expression in cells, could catapult the drug discovery method by identifying drug targets from a large database of drugs (12).

Deep learning could also potentially be a tool for precision medicine and personalised treatment development (13). Deep learning models can integrate all patient-specific data, which includes gene profiling, clinical records, and personal information, with lifestyle identification, which allows for the development of precision medication and treatments curated for the specific individual (14). Deep learning is also capable of understanding and analysing large amounts of data with high accuracy, allowing it to identify various genetic markers, genetic variations, biological markers, drug efficacy, protein interactions, and clinical prognosis, which are interconnected with identifying the best treatment responses and disease progression identification (15). For example, Dinov et al. (16) developed a machine learning protocol based on data processing and characterization for Parkinson's disease diagnosis that incorporated various learning models that achieved high accuracy in detecting these diseases in patients and could be utilised to enhance deep learning for drug discovery and personalised treatment based on the same principle.

Deep learning models could handle large and complex datasets. Deep learning algorithms can analyse and process various complex biological or genomic data at ease, which could be done by extracting intricate patterns and learning the best way to archive or select related data, which enhances their accuracy in predictions and classification of data (8). With the capacity to learn new things more precisely, deep learning model algorithms are also capable of extracting relevant features based on their prediction capacity from huge volumes of data, which could reduce the need for human intervention (17). For example, this is beneficial when it comes to the fields of biomedicine and molecular biology, which often come with complex and heterogeneous data that could pose a certain challenge in navigating them in a certain amount of time.

The scalability and transferability of deep learning models make them better at handling large volumes of complex data. Deep learning models can be trained by inputting one biological data task at a time with basic modifications, which over time could reduce the number of resources needed to train the model and help to improve the generalisation of new data (18). In addition, deep learning methods possess the capacity to identify novel patterns or relationships within data that may be difficult to discern using conventional analytical approaches. For instance, Romo et al. (19) demonstrated by training three different deep learning networks with three signal models consisting of tZ, vector-like quarks, and heavy glucan decay that their model is able to detect new physic signals with better accuracy.

### 2.2 Challenges of Using Deep Learning

One of the most common challenges when using deep learning for computational biology is interpretability (11). Some models of deep learning are sometimes regarded as "black boxes" due to their complexity, architecture, and internal representation, which could make it difficult to understand how the model could correctly predict biological mechanisms. In addition, ensuring interpretability is crucial for establishing trust among its stakeholders. When it comes to medical diagnoses, it is essential to verify that the model's decisions are based on reliable factors rather than being influenced by data artefacts. Hence, some efforts are needed to develop interpretability techniques that can shed light on the decision-making process of deep learning models (8).

Although deep learning could help the medical field produce almost accurate diagnostic values, it could also raise some ethical and regulatory issues. The utilisation of deep learning in the medical field could potentially raise ethical and regulatory issues involving patient privacy protection, which could be mitigated by producing or enhancing a current guideline that could include informed consent and data protection (20). In addition to that, deep learning models could also produce a biassed diagnostic report, which could conflict with some ethical values. This could pose a risk in that the deep learning model will produce a different prediction that will discriminate between different groups of patients, which could lead to wrong diagnostic results or potentially deny them access to treatment (21, 20). Thus, transparent and proper usage of deep learning models ensures accountability in the field of biomedical research and healthcare without discrimination.

However, despite being capable of retrieving a large volume of datasets at once, deep learning models could only read and interpret the quality and labelled datasets for functional training. This could pose a difficulty in acquiring a high-quality data set, especially in the fields of healthcare and biomedicine, due to the limited data provided, which could be constrained by privacy protection and regulations in addition to data heterogeneity (22). In addition to that, in order to acquire biological or biomedical data, it frequently comes from multiple sources, such as electronic health records and pathological data, with different formats and standards that could hamper the performance and generalisation of the deep learning model.

Despite the state-of-the-art features of the deep learning model, it will require an extensive computational component and often significant resources to be trained and utilised effectively by its user. Deep learning models require a high-performance computing infrastructure and specialised hardware that are capable of training large-scale deep learning models. This could pose such a challenge, especially when it comes to small non-profit organisations or research institutions with limited resources and infrastructure in place (23).

## 3.0 DEEP LEARNING MODELS FOR DNA SEQUENCE CLASSIFICATION AND PREDICTION

The utilisation of deep learning models has gained significant popularity in analysing and categorising DNA sequences, facilitating scientists making precise forecasts about the structure and function of genetic matter. The intricate patterns found in genomic data can be comprehensively analysed using artificial neural network-based algorithms employed by deep learning technology, making it ideal for recognising important genetic attributes while predicting with accuracy the health implications associated with hereditary variations that cause diseases.

### 3.1. Role of Deep Learning in Genomic Variant Detection and Precision Medicine

The application of deep learning, a subcategory within machine learning, has revolutionised genomic variant detection and gene expression analysis. Genomics encompasses an organism's entire genetic makeup and provides vital information that aids in understanding biological processes, illnesses, and individual variances. By utilising the potential of deep neural networks, significant advancements have been made by researchers in comprehending gene expression profiles and genetic variations. This has led to valuable insights related to personalised medicine, drug discovery, and the mechanisms underlying diseases. For genomic variant detection specifically, these algorithms play a crucial role by precisely classifying variants leading up to identifying disease-causing mutations (24). Simultaneously, for gene expression analysis purposes, studies can be done on splicing-code model interpretation or identification of long noncoding RNAs (24).

The use of deep learning in genomic variant detection has enabled the prediction of the organisation and functionality of various genomic elements such as promoters, enhancers, and gene expression levels (25). Deep learning is utilised in the detection of gene variants to anticipate their effect on disease risk and gene expression. To accomplish this, a genome is split into optimal, non-overlapping fragments using fragmentation and windowing techniques (26). A three-step procedure—fragmenting, model training for forecasting variant effects, and evaluating with test data—constitutes deep learning-based identification of genetic variations (26). A deep learning model demonstrated favourable precision in distinguishing patients from controls and the ability to identify individuals with multiple disorders during research on genetic variants in non-coding areas (27). The genomic regions that carried the most significant weights of genes were seen to be enriched with biological pathways related to immune responses, antigen/nucleic acid binding, the chemokine signalling pathway, and G-protein receptor activities. These results could aid future investigations regarding mental illnesses by providing mechanistic understandings through supporting genetic markers (27).

Additionally, deep learning methods, like convolutional neural networks (CNNs), have been utilised to forecast genetic variations that could potentially cause illnesses (25). In a particular study, a deep learning model was employed to forecast the functional implications of non-coding region genomic variations and proved more effective compared to conventional methods (25). Additionally, utilising deep learning capabilities also led to successful prognosis on gene expression levels influenced by single-nucleotide polymorphisms (SNPs). The advanced technology superseded traditional models while uncovering new SNPs linked to peculiarities in gene expression levels. (28)

The expression of genes is dependent on various transcriptional regulators such as pre-mRNA splicing, polyadenylation, and transcription for successful functional protein production. High-throughput screening technologies have provided valuable information on quantitatively regulating gene expression; however, experimental or computational techniques are unable to explore large biological sequence regions. Fortunately, deep learning has been successfully utilised for gene expression analysis, with several studies demonstrating its ability to predict levels and identify enhancer-promoter interactions. A deep learning architecture named Enformer was utilised in a study published in Nature Genetics (29) to enhance the accuracy of gene expression prediction from DNA sequences. The study found that by integrating data from long-range interactions up to 100 kb within the genome through massive parallel assays, predictions of saturation mutagenesis and natural genetic variants could be enhanced.

Moreover, gene expression analysis was explored through the utilisation of deep generative models (DGMs) in a study published by Springer (30). DGMs can identify underlying structures such as pathways or gene programmes from omics data and provide a representation by supplying a framework that accounts for both latent and observable variables. These models have proven effective in analysing high-dimensional SNP data by facilitating joint analyses across multiple loci to understand multigenic diseases. In addition, researchers are using deep learning approaches with DGMs on high-dimensional SNP data to predict how nucleotide changes will affect DNA levels beyond just genetic expression datasets alone.

The utilisation of deep learning in the identification of genomic variants and analysis of gene expression has revolutionised our comprehension of genetics, opening doors for developments in individualised medical treatments. By accurately pinpointing genetic variations and interpreting gene expressions, this technology hastens the discovery process for disease-associated genes, drug targets, and potential therapeutic interventions. Additionally, computational tools enabled by deep learning allow clinicians to make refined decisions based on individual genomic profiles. Although there are limitations such as overfitting and interpretability issues that accompany deep-learning-derived techniques despite their promising results so far, traditional methods have been outperformed under certain circumstances. Luckily, a multitude of pipelines are available, facilitating their utility within genomics research.

### 3.2. Advancements in Deep Learning for Epigenetic Data Analysis

In recent times, there have been significant strides made in utilising deep learning methodologies to parse epigenetic data. These advancements have revolutionised our comprehension of the intricate regulatory mechanisms that control gene expression and chromatin dynamics. The study of changes occurring within genes' expression without any alteration to their underlying DNA sequence is known as epigenetics, which plays a critical role in various biological processes and ailments. Genetic factors, along with environmental influences such as nutrition and lifestyle, can affect these modifications found within an individual's genetic make-up. Deep learning techniques provide powerful tools for extracting key insights from this information, particularly regarding obesity or metabolic conditions (31).

A significant advancement observed is the usage of convolutional neural networks (CNNs) to examine DNA methylation designs. The modification, a widely researched epigenetic process known as DNA methylation, holds considerable importance in controlling gene expression. With their ability to capture spatial dependencies and patterns seen in such data, CNNs hold great promise for this field. For example, DeepCpG, an innovative model developed by Angermueller et al. (2017), utilises CNNs that are trained on both the relevant attributes within sequences as well as methylation pattern recognition abilities, leading to accurate predictions regarding changes with respect to DNA's methylated state at different locations throughout the genome. Its superior results compared to traditional methods mean that it can provide more precise analyses than previous systems used until now across entire genomes (32).

Epigenetic alterations possess crucial repercussions on an individual's health that are biassed by environmental aspects (e.g., exercise, stress, and diet). With recent advancements in deep learning-driven approaches, it has become feasible to swiftly scrutinise substantial volumes of epigenetic information. One profound application is the development of DNA methylation ageing clocks that leverage molecular-level features to accurately predict age based on large-scale datasets (34). DeepMAge is an example of a DNA ageing clock that was created through training a deep neural network using 4,930 blood DNA profiles from 17 research studies. It has proven to have better results than other ageing clocks that rely on linear regression methods, as evidenced by its independent verification set, which contains 1,293 samples sourced from 15 additional studies and exhibits an absolute median error rate of only 2.77 years (34).

Additionally, the analysis of histone modification data has been explored using deep learning techniques. Key markers for gene activity and chromatin structure include various modifications such as acetylation and methylation. To unravel the intricate connection between patterns in these modifications and gene expression, neural networks like attention-based ones or those based on deep belief have proven effective. In particular, Yin (2019) introduced their model called

DeepHistone, which leverages multiple profiles from different histones to predict levels of gene expression with high precision, leading to new insights into epigenetic mechanisms previously unknown (33).

Moreover, studies conducted on animals have shown that epigenetic modifications are linked to metabolic health outcomes in humans. Animal models provide ideal opportunities for rigorously controlled studies that can offer insight into the roles of specific epigenetic marks in indicating present metabolic conditions and predicting future risks of obesity and metabolic diseases (31). Examples include maternal nutritional supplementation, undernutrition, or overnutrition during pregnancy, resulting in altered fat deposition and energy homeostasis among offspring. Corresponding changes in DNA methylation, histone post-translational alterations, and gene expression were observed, primarily affecting genes regulating insulin signalling and fatty acid metabolism (31). Recent studies indicate paternal nutrition levels also affect their children's fat disposition, with corresponding detrimental effects on their bodies' epigenetic characterizations (31).

Although deep learning-based techniques demonstrate potential in epigenetic data analysis, they possess constraints. Substantial amounts of top-notch data are necessary for these models to train adequately. Additionally, interpreting results from deep learning can be challenging; thus, understanding biological mechanisms leading to model predictions is difficult. Henceforth, evaluating the quality of input and model performance prior to significant endorsement of outcomes becomes critical.

The latest advancements underscore the promise of deep learning methods for scrutinising epigenetic data. Neural networks' potency allows scientists to discern concealed patterns, grasp far-reaching relationships, and make precise forecasts from extensive epigenomic datasets. These progressions offer significant enlightenment into gene expression's regulatory mechanisms, which can aid in comprehending diseases and designing specific treatments. The initiatives undertaken by these experts are merely a few illustrations of the thrilling headway attained within this domain, sparking further innovations in research on epigenetics.

### 3.3 Applications of deep learning in Protein Structure Prediction

The prediction of protein structure has been notably affected by the utilization of deep learning. With its precise and efficient methods, it has transformed this field entirely by providing a powerful means for determining the three-dimensional shape of proteins. Predicting their structure plays an important role in comprehending their function, drug discovery, and designing therapeutics; thus, making it extremely crucial to accurately predict them. Effectively capturing complicated patterns and dependencies within protein sequences leading to predicting their corresponding structures successfully has marked remarkable success using deep learning models.

Predicting the structure of a protein with precision, based solely on its sequence, proves to be challenging, but deep learning presents itself as a viable solution. Recent applications employing this approach have successfully predicted both three-state and eight-state secondary structures in proteins (35). Protein secondary structure prediction serves as an intermediate process, linking the primary sequence and tertiary structure predictions. The three traditional classifications of secondary structures include helix, strand, and coil. However, predicting 8-state secondary structures from protein sequences is a much more intricate task referred to as the Q8 problem- which offers greater precision in providing structural information for varied applications. Thus, several techniques of deep learning such as SC-GSN network, bidirectional long short-term memory (BLSTM) approach, a conditional neural field with multiple layers, and DCRNN have been employed to forecast the eight-state secondary structures (35). In addition, a next step conditioned convolutional neural network (CNN) was utilized to identify sequence motifs linked with particular secondary structure elements by analyzing the amino acid sequences. For instance, in 2019, AlQuraishi's research brought forth "Alphafold," a CNN-powered model that accurately forecasted protein secondary structure. Its competence in capturing sequence-structure connections resulted in better forecasts when weighed against conventional means (36).

Besides that, the significance of deep learning in forecasting protein-protein interactions and binding sites cannot be overstated. Through a thorough examination of protein sequences and structures, profound machine models can detect intricate connections between proteins while pinpointing the regions that interact with ligands or other proteins. In Xiuquan Du’s (2017) study, they developed an ingenious "DeepPPI" model which effectively predicts interactions among different types of proteins via their sequence information. The keen ability to extract critical elements present within these constituents enables DeepPPI to identify probable partners for interplay; thereby increasing our comprehension of how protein constructs form as well as function all at once. (37)

Significant advancements have been made in the tertiary structure prediction of proteins using deep learning. Abriata et al., employed a deep learning contact-map approach to achieve a notable breakthrough in the 13th Critical Assessment of Techniques for Protein Structure Prediction (CASP13) (38). To determine protein folding accurately, predicting residue-residue contacts is crucial. Deep learning approaches leverage vast protein databases to capture intricate patterns and dependencies between residues. This aids in long-range contact prediction by developing deep-learning models that guide the assembly of protein structures with greater precision. Wang et al.'s (2021) method utilized a deep residual network which proved effective in anticipating residue-residue interactions for precise folding predictions through their model's accuracy improvement (39). Meanwhile, the "AlphaFold 2" model created by Senior et al. (2020) is another significant illustration worth noting. Through the integration of RNNs and attention mechanisms, AlphaFold 2 achieved extraordinary precision in prognosticating protein tertiary structures, surpassing other techniques in the Critical Assessment of Structure Prediction (CASP) competition as well. Such success can be attributed to how RNNs effortlessly capture long-range dependencies within protein sequences without issue. (40)

These applications showcased the extensive range and influence of deep learning in predicting protein structure. With its adeptness at identifying complex patterns and connections within protein sequences and structures, deep learning has enabled significant progress in comprehending aspects such as folding, function, and interactions related to proteins. Although there may be upcoming challenges and opportunities, the extensive implications of deep learning's capability to reveal fresh insights regarding proteins are immense in terms of comprehending basic life processes, personalized medicine, as well as drug discovery.

## 4.0 DEEP LEARNING MODELS FOR PREDICTING PROTEIN STRUCTURE FROM SEQUENCE DATA

Deep learning refers to a type of machine learning based on a neural network that utilizes multiple levels of processing to extract results progressively (41). These methods have dramatically improved. A Myriad of researchers had initially employed machine learning (ML) methods to overcome problems and aid in research that is in the field of computational biology such as protein prediction and protein interaction. Deep learning algorithms have the ability to control huge complex and raw data as well as learn beneficial and abstract functions. Deep learning has also been utilized in the field of bioinformatics on tasks such as increasing the amount of data (42,43). In general, deep learning architecture may accept a diverse range of input sources of data for analysis for downstream purposes such as the 3D structure of the protein, network topology, domain components, primary sequence, text mining, and many more. Moreover, Neural network modules, which are very important for deep learning, traditionally include a few types namely a fully connected layer, convolutional layer, recurrent layer, and many more (44,45).

### 4.1 Applications of deep learning in Protein-protein interaction Prediction and drug discovery

The latest deep learning techniques that are employed in PPI models may include Deep convolutional neural networks. This technique is widely used due to its potential to extract features from structural data. For instance, based on Torrisi et al (44), the structural network information along with the sequence-based features predicts the interactions between proteins. Besides that, in order to extract structural information from 2D volumetric representations of proteins, the pre-trained ResNet50 model was used. The results indicate that methodologies for image-related tasks can be extended to work on protein structures (45). However, these techniques of analyzing molecular structure have drawbacks such as elevated computational expenses and as well as interpretability (45).

There are various deep learning methods that could be utilized for protein-protein interaction networks. First, the DeepPPI is a multilayer perception learning structure that requires protein sequences as its source of input features (46). The encoding method utilized by this method is the seven sequence-based features which use concatenation as its combining method. Moving on to The second method which is DPPI, is a convolutional neural network structure that also uses protein sequences as its source of input features (47). protein-positioning specific scoring matrices, PSSM which is derived by PSI-BLAST is used as the encoding method for this deep earning method. Next, the DeePFE-PPI is also a method that was created in 2019 using multilayer prescription which uses protein sequences as an input. The encoding method that is utilized in this method is pre-trained model embedding (Word2vec) (48). Besides that, S-VGAE is also an example of graph convolutional Neural networks which utilized protein sequences and topology information of protein-protein interaction networks. The encoding method employed in this technique is a conjoint method and it is combined via the concatenation method (49).

Besides protein-protein interaction, Deep learning is also utilized in drug discovery for optimizing the properties of drugs, determining new drugs as well as predicting drug-target interactions. In addition, deep learning is also employed in predicting the molecular properties of drugs such as solubility, bioactivity, toxicity, and many more (50). In addition, it is also used to produce novel molecules that have preferred properties. Next, in QSAR studies for drug discovery, the deep neural network is used to predict the bioactivity of the drugs and their chemical structures (50).

Moreover, deep learning methods are also applied to leap the optimization of incorporating traditional in silico drug discovery efforts. Based on the research which is entitled, (AtomNet from Atomwise company, the first major application of deep learning into DTI prediction) clearly shows the application of convolutional neural networks which is a type of deep learning technique to predict the molecular bioactivity in proteins (51). In addition, in terms of docking, deep learning techniques have been employed to improve the accuracy of both traditional docking modules and scoring functions. For instance, the docking proved that the application of deep learning had improved the binding mode prediction accuracy over the baseline docking process. Besides that, this paper had also proven the fact that Deep learning could be successfully utilized in the rational docking process (52).

### 4.2. Recent developments in deep learning-based techniques for analyzing protein function and evolution.

Recently, there are a few developments that are used in protein analysis by incorporating deep learning algorithms. An example of it would be combining deep learning with homology modelling.  Furthermore, Homology modelling is the most popular protein structure prediction method that is utilized to generate the 3D structure of a protein. This is done based on two principles which are the amino acid that is used to determine the 3D structure and the 3D structure that is preserved with regard to the primary structure (53). Therefore, it is convenient and a effective way to build a 3D model using known structures of homologous proteins that have a certain sequence similarity. However, it does have some challenges when using this method such as weak sequence structures, modelling of the rigid body shifts and many more (53). However, incorporating deep learning models has resulted in great improvement in the protein’s model accuracy.

The deep learning-based methods are employed to improve accuracy in each step of template-based modelling of protein. For instance, DLPAlign is an example of a deep learning technique that is combined with sequence alignment (54). This straightforward and beneficial approach may aid to increase the accuracy of the progressive multiple sequence analysis method by basically providing training to the model based on convolutional neural networks CNNs (54). Besides that, DESTINI is also a recent method which applies deep learning techniques algorithm, for protein residue and residue contact prediction along with template-based structure modelling (55).

In short, Deep Learning techniques have provided various achievements in collaborative sectors, namely model quality assessment (QA), a subsequent stage in protein sture prediction. Basically, QA is followed by structure predictions to quantify the deviation from the natively folded protein structures in both template-based and template-free techniques.

### 4.3. Challenges and future directions

There are various challenges when using Deep learning techniques when analyzing biological-related specimens such as protein structure prediction. First, deep learning requires a large amount of high-quality data. Hence, only biological analysis could be done if only a large amount of data is gathered (56). Next, the deep learning model is incapable of multitasking when it is applied in an analysis procedure. Deep learning models are capable of handling one issue at a time. Furthermore, the interpretability of deep learning models is also a challenge of interest for many researchers to overcome. This is because it is difficult to understand and identify how they obtain their predictions. New techniques are being developed by researchers to overcome this problem. The future direction of deep learning is to create hybrid models by incorporating other machine learning techniques to improve performance and interpretability (56).

## 5.0 KEY CHALLENGES IN APPLYING DEEP LEARNING TO BIOLOGICAL DATA

As mentioned in the previous section, high-quality data, the inability to multi-task and data interpretability are some of the key challenges experienced in the application of AI systems such as deep learning into biological data. There are several other challenges, especially in terms of ethics and social implications which are addressed in the sub-sections below. Addressing these challenges of deep learning requires specific and innovative approaches specific to the types of biological data used. Thus, overcoming these challenges would ultimately pave a path to improvement in biological research.

### 5.1 Emerging Areas of Research and potential applications of deep learning in computational biology

Computational biology is defined as an interdisciplinary field which involves the use of techniques from various other fields such as biology, mathematics, statistics, computer science and more. Applications of deep learning in computational biology can be seen in various areas including in the study of genomics and proteomics. There are many major achievements that are obtained specifically in areas such as protein structure prediction, and rapid advancement in other areas of research from the traditional approaches including genomic engineering, multi-omics, and phylogenetics can also be seen. (57)

The study of genomes and their interaction with other genes and external factors is commonly known as genomics. One of the primary studies conducted in genomics is the study of regulatory mechanisms and non-coding transcription factors (58) One of the major current applications of deep learning research of genomics and transcriptomics is one of the emerging areas of research in deep learning. Deep learning is used to identify variations in genomic data, this includes DNA sequencing and gene expression. For example, it is used to predict the functions of genes, discover gene regulatory networks, and identify biomarkers in diseases. As a result of this application, the metabolic pathways can also be optimized. A study identified several challenges in genomics including mapping the effects of mutation within a population and the DNA sequence prediction in a genome which has complex interactions and variations. To combat these challenges, deep learning methods are employed in genomic studies. Deep Neural Networks (DNN) and Convolutional Neural Networks (CNN) are algorithms of deep learning that are employed in genomic studies (59). The DNN is able to solve the DNA sequence prediction problem as the model is trained on DNA sequence datasets and the corresponding protein structures. This enables the identification of the proteins which are specific and binds to a particular DNA sequence. The DNN models are also able to predict splicing outcomes for new DNA sequences based on the training of splicing patterns. CNN, on the other hand, addresses the remaining issue; the prediction of mutation effects (58). This model is able to analyze and identify the potential causes of mutation in a DNA sequence and to determine the mutation or disease on the single nucleotide variant that is affected. Both CNN and DNN are powerful tools of deep learning which can provide valuable information on the complex structures of genomes. The application of the algorithm in the field of genomics would greatly improve the analysis of complex structures, functions and interactions of genomes.

In the field of biological image analysis, the deep learning algorithm CNN is found to be an efficient tool that is able to undertake several tasks such as classification, feature detection, pattern recognition and feature extraction (58). Since the CNN models are effective in processing grid-like data such as images, it is commonly utilized in image analysis (59). Staking more convolutional layers in the model aids in detecting complex and abstract features in biological images. The CNN model is able to learn and identify delicate patterns and subtle differences in biological images which improves the accuracy of a diagnosis. DeLTA is an example of a deep learning tool used to analyze biological images, specifically, time-lapse microscopy images (60). The Deep Learning for Time-lapse Analysis (DeLTA) is able to analyze the growth of a single cell and the gene expressions in microscopy images. It was found to be able to process and capture microscopy images with high accuracy and without the need for human interventions. Furthermore, deep learning is incorporated into healthcare, specifically radiology. Tasks such as classifying patients based on chest X-rays diagnosis and nodule detection in computed tomography images are done using deep learning. (61) The analysis of a large number of radiology data depends on the efficiency of the powerful deep learning algorithms.Thus, deep learning algorithms have the potential to revolutionize biological image analysis by providing automated and accurate analysis.

Deep learning in the proteomics field can mainly be shown in the protein-protein interactions predictions. Protein complexes can also be identified through deep learning. Protein structure and function can then be predicted from the data obtained and be used for various activities such as the identification of targeted proteins for drug development. Deep learning is applied in the study of phylogenetics where the limitation in the classification methods where the branch lengths of the phylogeny cannot be inferred is to be overcome. (57)

While there are many applications of deep learning that bring significant advancements, there are other potential applications of deep learning in the biological field that can be further discussed and implied. For instance, deep learning is applied in the identification of protein-protein interactions. Therefore, a similar technique can be applied in drug design. In terms of drug design, the application and incorporation of deep learning have made the process more time and cost-efficient as compared to traditional drug design methods (62). The use of deep learning in drug design is identified to be more flexible due to the neural network architecture of the algorithm (63). Especially in the current era, with the combat against COVID-19, deep learning has shown great potential in accelerating the drug design process. The deep learning models are able to identify antimicrobial compounds against a disease or a virus by training the model with the ability to identify molecules against the virus or bacteria. Similarly, another study showed the use of deep learning models in the use of de novo drug design where the model was trained to identify the physical and chemical properties of the drugs, classifying them based on their features and allowing automated extraction to create a novel ligand against the target protein (65). Drug repurposing, which is a quicker method to achieve and complete drug designing for a disease or illness, is found to incorporate deep learning approaches. An article reported the use of network-based approaches in drug repurposing to identify the target molecule for known drugs to speed up the process (66). Another study in relation to COVID-19, used the Molecule Transformer-Drug Target Interaction (MT-DTI), a deep learning model trained with chemical sequences and amino acids sequences to identify the commercially available antiviral drugs with similar properties of interaction with the SARS-CoV-2 virus (64). These are just some examples of the emerging use of deep learning models in drug design. The appearance of the COVID-19 disease has boosted the application of AI systems in the field to improve the speed and efficiency of the process.

To summarize, the deep learning algorithm is a powerful artificial intelligence tool that is widely used in the field of computational biology. The application of the tool is just in its beginning phase as there are more fields and complex challenges are to be explored and tackled in the upcoming future. The application of this AI technology will help to shape the future of computational biology by improving the predictions and the understanding of biological processes.

### 5.2. Ethical and social implications of using deep learning in biological research.

The advancements in digital technology allow the incorporation of Artificial Intelligence tools such as Machine Learning and Deep Learning in various fields of research. These techniques use various algorithms to identify complex and non-linear correlations in massive datasets and have the ability to improve prediction accuracy by learning from minor algorithmic errors encountered. Despite the use of a powerful machine learning tool, such as deep learning algorithms, in the field of research and healthcare is found to be revolutionary, it inevitably raises ethical concerns and social implications that require careful consideration (67).

Four major ethical issues were identified in regard to the use of AI tools in the healthcare system; informed consent of data, safety and transparency, algorithmic fairness and biases and data privacy (68). These concerns may be identified in the healthcare sector but these concerns are also integrated in the usage of deep learning techniques in biological research which involves the use of deep neural networks to analyze and interpret volumes of biological data.

The field of biological research involves the use of various biological data which includes genomic sequences, protein structures, medical images and scans and more. The deep learning algorithm has access to these data, containing sensitive biological data, to aid in research. As medical records and genomic information of individuals are involved and used, it is critical to ensure the privacy of the individual, as well as the informed consent for data usage, is obtained (68). Privacy violations and mishandling of personal information are examples of invasion of data privacy without individual consent (70). Safety protocols are to be established to protect data confidentiality and to obtain the appropriate consent from the participants.

The potential bias in the algorithms of the deep learning tool is one of the main ethical concerns surrounding AI systems. The algorithms utilized in these systems can perpetuate biases and negatively impact marginalized groups (68). This is because the training data is not representative of the diverse populations leading to biased results and disparities in research outcomes. The biases can be found in different stages of biological research, including data collection and annotation, if they are not addressed (71). Therefore, efforts should be taken to address the bias in data collection to promote the inclusivity of all data regardless of population type, disease groups, diversity and other factors involved in research.

It is also crucial to promote transparency and safety in the AI tools used in biological research. Deep learning models are considered to be opaque making it difficult to understand the process of the prediction and decision in research. As deep learning models are made up of multiple layers of artificial neurons, where each layer corresponds to a different learning pattern, it poses a challenge to accurately identify the pattern learned by each layer functions to make a prediction. Lacking transparency in the algorithm decision-making process begins the questioning of the ability to scrutinize the AI results as a reasonable explanation leading to the data being unprovable and uninterpretable by humans (68, 71). Thus, transparent models are required to make sure the researchers are able to observe the prediction pattern to validate and understand the results obtained. This ensures the accountability of the scientific research process. However, it was discussed that full transparency may cause friction against certain ethical concerns as it may leak private and sensitive data into the open (69). Hence, there should be limitations to the disclosure of the algorithms.

Security in biological research not only involves maintaining the data and privacy of personal data, but it also involves the responsible use of technology. The technology at hand, deep learning, must be used responsibly and ethically in research. Scientists must incorporate ethical frameworks to avoid potential misuse and unintended consequences or risks that may occur (68). Furthermore, risk assessments are to be conducted to aid in decision-making and reduce the possible negative impacts. The deployment and implementation of this technology must be considered well with proper safety measures, regular monitoring and evaluations.

Moreover, there is the concern of liability and accountability where questions would arise to who would be the person to be held accountable towards any form of mistakes or errors caused by deep learning algorithms used in research (72). As the algorithms are continually learning and evolving, it is difficult and complex to determine liability. Thus, legal frameworks are to be adapted with clear lines of responsibility to address the challenges faced by the AI system in biological research (73).

Ensuring equitable access to the deep learning tool is one of the social implications of AI systems in biological research. Promoting equitable access to scientists and researchers would be able to participate in the advancement of deep learning algorithms in biological research (74). It would also prevent exacerbating disparities in biological research while promoting an inclusive and collaborative research environment. Fostering the exchange of ideas and knowledge of experts would further aid deep learning to be integrated into the biological research community.

In a nutshell, these are some of the social implications and ethical concerns revolving around the use of AI systems such as Deep Learning in the field of biological research. Deep learning is the future of more efficient and advanced research; however, the ethical concern mentioned above should be addressed to ensure the responsibility and accuracy of the algorithm in biological research.

## 6.0 SUMMARY OF RECENT ADVANCES IN DEEP LEARNING FOR COMPUTATIONAL BIOLOGY

### 6.1 Future Prospects and potential impact of deep learning on biological research and clinical practice

The application of deep learning models is poised to have a substantial influence in the field of biological research and clinical practices. With their ability to analyze large and intricate data, deep learning models could be beneficial in assisting for pathological diagnosis, drug discovery, genomic data identification or even personalized treatments. By harnessing deep learning algorithms researchers can examine biological data consisting of gene expression or protein structure to identify new patterns or molecules which could yield an insight into the biological structure mechanisms. In addition to that, deep learning models are also being used to facilitate accelerating new drug target development and research, developing new accurate diagnostic tests, and aiding in improving clinical trial designs (75). The future prospect of deep learning models in both biological research and clinical practice are promising given that this technology and its algorithms continue to be developed which could potentially catapult humanity into a new era of making diagnosis for diseases or illnesses in addition to providing a more practical way for providing a better patient care with precise treatments and prevention of diseases (76). On top of that, deep learning models have the potential to be used to improve the efficiency and effectiveness of healthcare delivery systems by automating menial tasks and accelerated diagnostics tests (77). Deep learning models possess the ability to integrate and analyze a variety of data types, including genomics, proteomics, imaging, and clinical data. This enables the exploration of concealed patterns and relationships within these datasets, empowering machine learning to offer a holistic comprehension of diseases and provide guidance for translational research endeavors. Besides that, deep learning has broad applicability in addressing diverse challenges. By training on extensive datasets, deep learning models excel at navigating tasks such as image classification, object detection, speech recognition, and machine translation aside from that, deep learning is a rapidly growing field, and it is being used in a variety of domains, including healthcare, computer vision, natural language processing, and robotics (78). The summary of the recent advancements of deep learning in computational biology can be referred to **Table 2.**

**Table 1** The table depicts a historical and evolutionary overview of deep learning.

| | | |
|---|---|---|
| 1873 | A. Bain | The earliest models of neural networks, called Neural Groupings, were introduced and were inspired by the Hebbian Learning Rule. |
| 1943 | McCulloch & Pitts | The MCP Model was introduced, which is considered the precursor to Artificial Neural Models. |
| 1949 | D. Hebb | Considered as the father of neural networks, he introduced the Hebbian Learning Rule, which formed the basis for modern neural networks. |
| 1958 | F. Rosenblatt | The first perceptron, which closely resembles modern perceptrons, was introduced. |
| 1969 | Minsky and Papert | Publish Perceptrons, which criticizes the perceptron and limits the potential of neural networks |
| 1974 | P. Werbos | Introduced Backpropagation |
| 1980 | T. Kohonen<br><br>K. Fukushima | Introduced Self Organizing Map<br>Neocogitron was introduced, which served as inspiration for Convolutional Neural Networks. |
| 1982 | J. Hopheld | The Hopfield Network was introduced |
| 1985 | Hilton & Sejnowski | The Hopfield Network was introduced |
| 1986 | P.Smolensky<br><br><br>M. I. Jordan | Introduced Harmonium, which is later known as Restricted Boltzmann Machine<br><br>Defined and introduced Recurrent Neural Network |
| 1990 | Y. LeCun | LeNet was introduced, demonstrating the practical potential of deep neural networks. |
| 1997 | Schuster & Paliwal<br><br>Hochreiter& Schmidhuber | Introduced Bidirectional Recurrent Neural Network<br><br>Introduced LS'TM, solved the problem of vanishing gradient in recurrent neural networks |
| 2006 | G. Hinton | Deep Belief Networks were introduced, along with the layer-wise pretraining technique, which marked the beginning of the current deep learning era. |
| 2009 | Salakhutdinov& Hinton | Deep Boltzmann Machines were introduced. |
| 2012 | G. Hinton | Dropout, an efficient method for training neural networks, was introduced. |
| 2012 | Alex Krizhevsky, Ilya Sutskever, and Geoffrey Hinton | Convolutional neural network (CNN) for Image classification |
| 2014 | Ian Goodfellow, Yoshua Bengio, and Aaron Courville | Generative adversarial network (GAN) for image generation |
| 2020s | | Deep learning continues to evolve and is used for a wider range of tasks, including self-driving cars, medical diagnosis, and financial trading etc |

**Table 2** Recent developments in applying deep learning to computational biology. The table illustrates recent advancements in deep learning algorithms, their applications in computational biology, and relevant references.

.

| Deep Learning Algorithm | Application in Computational Biology | References |
|---|---|---|
| Convolutional Neural Networks (CNN) | Gene expression analysis | 79 |
| Recurrent Neural Networks (RNN) | DNA sequence analysis | 80 |
| Generative Adversarial Networks (GAN) | Synthetic biology and protein design | 81 |
| Deep Belief Networks (DBN) | Protein structure prediction | 82 |
| Reinforcement Learning (RL) | Drug discovery and optimization | 83 |
| Transformer Networks | RNA structure prediction | 84 |
| Autoencoders | Disease diagnosis and prognosis | 85 |
| Graph Neural Networks (GNN) | Protein-protein interaction prediction | 86 |
| Variational Autoencoders (VAE) | Single-cell genomics analysis | 87 |
| Deep Reinforcement Learning | Drug target identification | 88 |
| Capsule Networks | Protein structure classification | 89 |
| Adversarial Autoencoders | Gene expression imputation | 90 |
| Deep Boltzmann Machines (DBM) | Epigenetic data analysis | 91 |
| Deep Learning Algorithm | Application in Computational Biology | 92 |
| Attention Mechanism | Single-cell RNA sequencing analysis | 93 |
| Deep Q-Networks (DQN) | Drug toxicity prediction | 94 |
| Capsule Networks | Protein-protein interaction prediction | 95 |
| Deep Generative Models | DNA sequence generation | 96 |
| Graph Convolutional Networks (GCN) | Drug-target interaction prediction | 97 |
| Deep Survival Analysis | Cancer survival prediction | 98 |
| Transformer Networks | Transcriptomics analysis | 99 |
| Graph Neural Networks (GNN) | Drug repurposing | 100 |
| Adversarial Networks | Image-based phenotypic screening | 101 |
| Deep Transfer Learning | Drug response prediction | 102 |
| Generative Adversarial Networks (GAN) | Synthetic data generation | 103 |
| Deep Reinforcement Learning | Protein folding | 104 |
| Variational Graph Autoencoders (VGAE) | Disease-gene prioritization | 105 |
| Deep Neural Networks (DNN) | Metagenomic analysis | 106 |
| Convolutional Recurrent Neural Networks (CRNN) | Chromatin state prediction | 107 |
| Deep Clustering | Cell type identification | 108 |
| Deep Reinforcement Learning | Protein-ligand binding affinity prediction | 109 |
| Graph Convolutional Networks (GCN) | Drug response prediction | 110 |
| Long Short-Term Memory (LSTM) | RNA splicing prediction | 111 |
| Deep Reinforcement Learning | Antibiotic resistance prediction | 112 |
| Capsule Networks | Protein function prediction | 113 |
| Autoencoders | Single-cell epigenomics analysis | 114 |
| Deep Belief Networks (DBN) | Genetic variant classification | 115 |
| Transformer Networks | Protein-protein interaction network analysis | 116 |
| Graph Convolutional Networks (GCN) | Drug-target interaction network analysis | 117 |
| Recurrent Neural Networks (RNN) | Protein secondary structure prediction | 118 |
| Deep Reinforcement Learning | Gene regulatory network inference | 119 |
| Variational Autoencoders (VAE) | Metabolomics data analysis | 120 |
| Deep Belief Networks (DBN) | Drug side effect prediction | 121 |
| Capsule Networks | Cancer subtype classification | 122 |
| Convolutional Neural Networks (CNN) | Histopathology image analysis | 123 |
| Generative Adversarial Networks (GAN) | Synthetic biology and gene synthesis | 124 |
| Transformer Networks | Protein contact prediction | 125 |
| Deep Reinforcement Learning | Genome sequence assembly | 126 |
| Graph Neural Networks (GNN) | Cell type classification in single-cell transcriptomics | 127 |
| Autoencoders | DNA motif discovery | 128 |

**Figure 1** The bird's view of application of deep learning in computational biology.

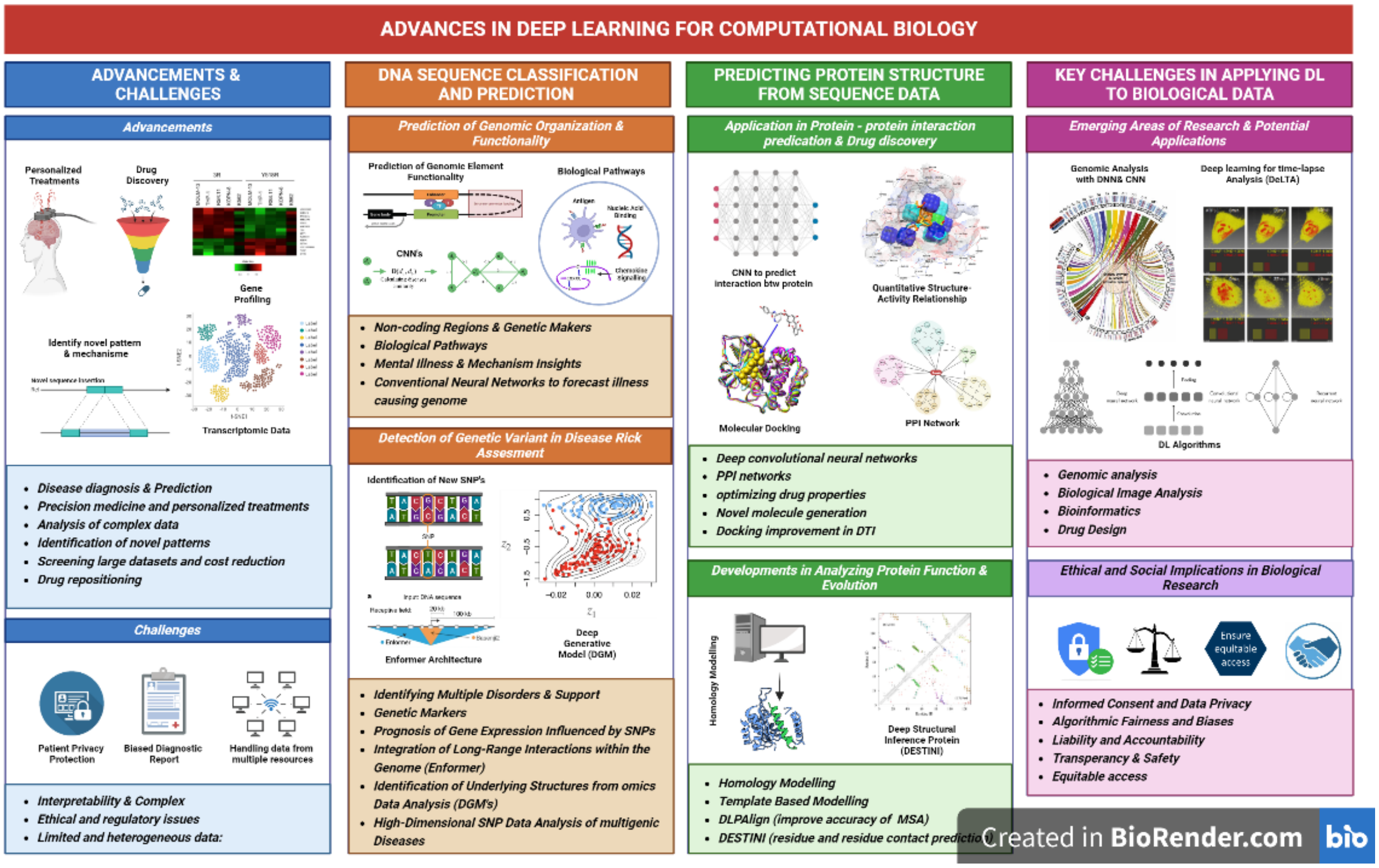